\documentclass[sigconf]{acmart}
\usepackage{graphicx}
\usepackage{subcaption}
\usepackage{xcolor}

\AtBeginDocument{%
  }

\copyrightyear{2023}
\acmYear{2023}
\setcopyright{acmlicensed}\acmConference[ICAIF '23]{4th ACM International Conference on AI in Finance}{November 27--29, 2023}{Brooklyn, NY, USA}
\acmBooktitle{4th ACM International Conference on AI in Finance (ICAIF '23), November 27--29, 2023, Brooklyn, NY, USA}
\acmPrice{15.00}
\acmDOI{10.1145/3604237.3626881}
\acmISBN{979-8-4007-0240-2/23/11}

\acmConference[ICAIF '23]{International Conference on AI in Finance}{November 28--29,
  2023}{New York, USA}

\begin{document}

\title{Deep Calibration of Market Simulations using Neural Density Estimators and Embedding Networks}


\author{Namid R. Stillman}
\email{namid@simudyne.com}
\affiliation{%
  \institution{Simudyne Limited}
  \country{United Kingdom}
}

\author{Rory Baggott}
\email{rory@simudyne.com}
\affiliation{%
  \institution{Simudyne Limited}
  \country{United Kingdom}
}

\author{Justin Lyon}
\email{justin@simudyne.com}
\affiliation{%
  \institution{Simudyne Limited}
  \country{United Kingdom}
}

\author{Jianfei Zhang}
\email{jianfeizhang@hkex.com.hk}
\affiliation{%
  \institution{Hong Kong Exchanges and Clearing Limited}
  \country{Hong Kong}
}

\author{Dingqiu Zhu}
\email{dingqiuzhu@hkex.com.hk}
\affiliation{%
  \institution{Hong Kong Exchanges and Clearing Limited}
  \country{Hong Kong}
}

\author{Tao Chen}
\email{taochen@hkex.com.hk}
\affiliation{%
  \institution{Hong Kong Exchanges and Clearing Limited}
  \country{Hong Kong}
}

\author{Perukrishnen Vytelingum}
\email{krishnen@simudyne.com}
\affiliation{%
  \institution{Simudyne Limited}
  \country{United Kingdom}
}

\acmArticleType{Research}
\keywords{Market simulator, Agent-based Models, Simulation-based inference, Neural density estimators, Embedding networks}

\begin{abstract}
The ability to construct a realistic simulator of financial exchanges, including reproducing the dynamics of the limit order book, can give insight into many counterfactual scenarios, such as a flash crash, a margin call, or changes in macroeconomic outlook. In recent years, agent-based models have been developed that reproduce many features of an exchange, as summarised by a set of stylised facts and statistics. However, the ability to calibrate simulators to a specific period of trading remains an open challenge. In this work, we develop a novel approach to the calibration of market simulators by leveraging recent advances in deep learning, specifically using neural density estimators and embedding networks. We demonstrate that our approach is able to correctly identify high probability parameter sets, both when applied to synthetic and historical data, and without reliance on manually selected or weighted ensembles of stylised facts.
\end{abstract}

\maketitle

\section{Introduction}\label{sec:intro}

\begin{figure}[!t]
\begin{center}
    \includegraphics[width=8cm]{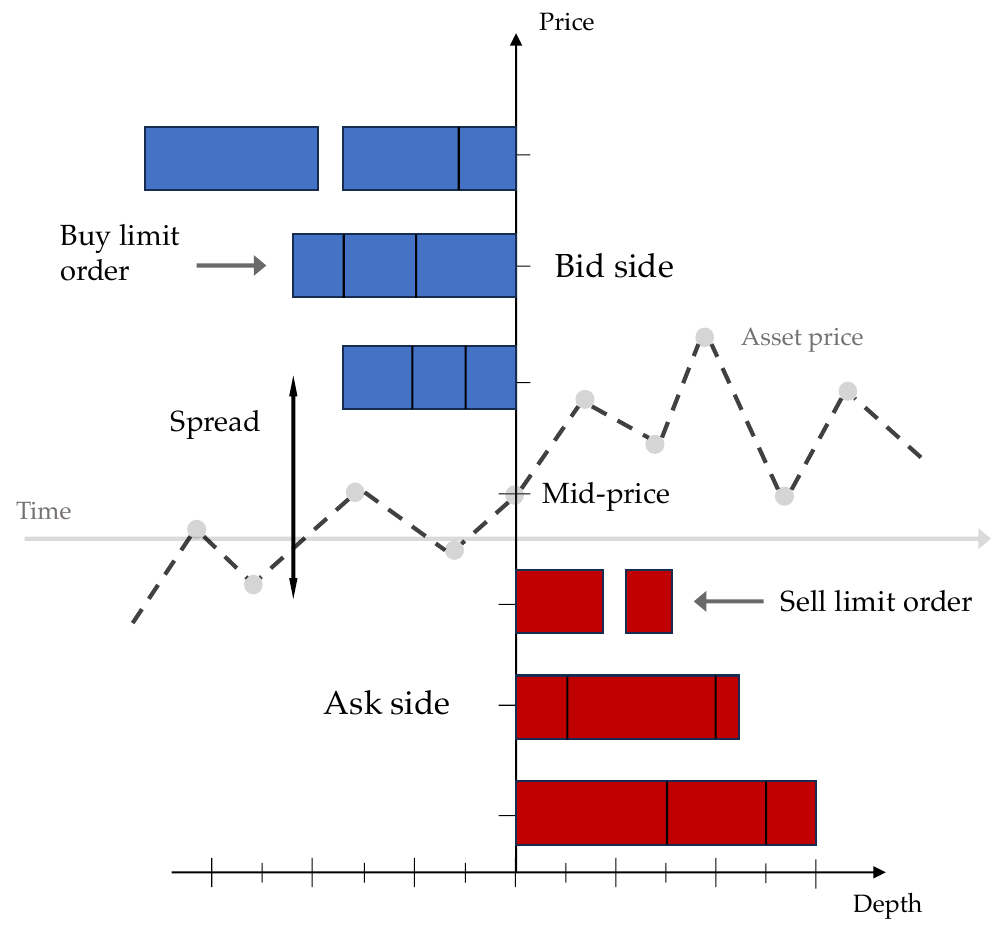}\qquad
\end{center}
\caption{Schematic demonstrating how a limit order book (LOB) handles different orders to buy or sell an asset. The spread marks the different between buy side and ask side, while the depth denotes the number of orders placed at a certain level of price. The mid-price or `touch' tracks the evolution of the asset price due to the exchange dynamics.}
\label{fig:LOB}
\end{figure}

Most major financial markets for equities, commodities and currencies, as well as other asset classes, operate on public exchanges where individuals place orders to buy or sell an asset. These markets are typically hosted on a centralised exchange which provides a platform for traders to place different types of orders for a security. Limit orders are orders to buy or sell at a limit price and are stored in a record kept by the exchange, also known as the limit order book (LOB) while market orders are orders to buy or sell a volume immediately from the LOB \cite{gould2013limit}. The LOB is updated each time an order is placed, amended, or cancelled. During a trading day, the market goes through a number of phases e.g., the pre-open call auction or the continuous double auction (CDA). Each phase has protocols that define which orders can be placed, how the orders are handled, when and how the LOB clears, and how those trades are priced. The underlying principle of the LOB is shown in \autoref{fig:LOB}. 

Securities traded on an exchange exhibit many of the characteristic signatures of financial time series, including being non-stationary, non-linear, and stochastic \cite{plerou2000econophysics}. Given the complexity of price dynamics, there is a strong need to explore the effect of counterfactual scenarios in order to minimise risk at the level of investor and exchange. These counterfactual scenarios include, for example, the market impact of an investor liquidating a very large position, a significant macroeconomic announcement such as labour market data, or a trading error such as a ``fat-finger mistake" \cite{easley2011microstructure}. 


Market simulators, which seek to reproduce the behaviours of the exchange, have been developed to better evaluate the impact of these and many other scenarios. These simulators aim to reproduce the underlying data-generating process and, hence, can be thought of as a form of generative modelling. Traditional methods have used empirical observations and domain knowledge to build models that incorporate approximations to trader behaviours in closed-form equations. These methods are often built as agent-based models (ABM), whereby a set of individual agents (traders) interact with one another according to a pre-defined set of rules (exchange protocol) \cite{samanidou2007agent,fagiolo2019validation}. We note that these models are typically structured such that the interaction network is many-to-one, i.e., all agents are connected to the exchange but not to each another. ABMs are appealing as they give an explicit means to both control and explain observations of market dynamics. For example, the strength in price fluctuations can be attributed to specific trader behaviours. Other methods, including those that rely on deep generative models, such as generative adversarial networks (GANs), are not as amenable to controlling for specific scenarios and typically rely on post-hoc analysis for explainability and control \cite{guzy2021evaluating,harkonen2020ganspace,cohen2021black}. 

While ABMs are typically easier to control and interpret than equivalent deep generative methods, they suffer from the so-called ``simulation-reality gap" of generative methods, which refers to the disconnect between simulated and historical data \cite{zhao2020towards,cruz2020closing}. This is typically due to the flexibility in model output which allows for a large variety of different scenarios to be generated but requires that parameters are appropriately constrained to best reproduce observations. Selecting the parameter sets that best match observations, also known as calibrating, remains a significant challenge to the deployment of market simulators as well as for the field of agent-based models more broadly. In this work, we present a novel approach for calibrating market simulators using simulation-based inference, which combines Bayesian inference with deep learning \cite{cranmer2020frontier}. We demonstrate that our method is able to infer parameters with high accuracy whilst also providing the entire posterior distribution over parameters, namely the probability distribution, $P(\theta|\mathbf{x})$, for a set of parameters, $\theta$, conditioned on the set of observations, $\mathbf{x}$. We use our method to calibrate two models of market simulations and infer parameters for historical data, demonstrating that we are able to reproduce many of the stylised facts observed in the data. 

\section{Relevant work}

Interest in the development of a realistic market simulator has increased in recent years due to new methods in deep learning and computer science as well as significant increases in data and compute. The history of market simulators can potentially be traced back as far as 1962, to the computer simulations of Nobel laureate George Stigler \cite{stigler1963public}. However, over the past several decades, both the number and the accuracy of market simulators has dramatically increased. For an overview of early examples of agent-based models of market simulations, we refer the reader to \cite{samanidou2007agent, axtell2022agent}. 

Of the many recent examples of ABM market simulators, one of the most popular and widely used is the Agent-Based Interactive Discrete Event Simulation (ABIDES) framework \cite{byrd2019abides}, which provides high-fidelity market simulations. Recent work has extended the ABIDES framework to include a reinforcement learning (RL) environment and a `hybrid' ABM/neural network-based market simulator \cite{amrouni2021abides,shi2023neural}. In this work, we use two custom-built ABMs for market simulation, which we describe in \autoref{sec:models}.

Separate to ABMs, deep learning has also been used to both recreate market behaviour. These approaches have used deep learning to reproduce market dynamics in an entirely data-driven way, without using explicit equations that describe market or trader behaviours. This includes using convolutional neural networks (CNNs) with recurrent architecture, in the so-called DeepLOB framework, as well as deep generative models such as GANs and variational auto-encoders (VAEs) \cite{zhang2019deeplob,coletta2021towards,buehler2020data}. These methods provide higher accuracy than traditional generative approaches but are more challenging to reproduce out-of-distribution or long-tail events such as those experienced during the COVID-19 pandemic or the global financial crises of 2008.

As discussed in \autoref{sec:intro}, traditional simulation approaches such as ABMs require parameter sets to be calibrated to best reflect observations. Methods for calibrating market simulators include using optimisation to find the point estimates of parameters, as well as Bayesian methods which seek to estimate the posterior over parameter values. Methods that use optimisation to find point estimates of parameter values include using Gaussian processes and surrogate modelling approaches that minimise the distance between historical and simulated data for a set of metrics \cite{bai2022efficient,gao2022understanding}. Other methods have used GANs to train a `calibration agent', that identifies real and synthetic data and uses this to choose parameters that are most likely to lead to realistic data generation \cite{storchan2021learning}.  

More broadly, methods that seek to estimate the posterior probability distribution of parameters are increasingly combined under the banner of {\it likelihood-free} or {\it simulation-based} inference \cite{cranmer2020frontier}. These methods include classical methods such as approximate Bayesian computation (ABC) as well as more recent advances that leverage deep neural networks such as neural posterior estimation (NPE). For a review of applications of simulation-based inference in the context of economics and financial timeseries, see \cite{dyer2022black}, which demonstrates how these methods can be applied to models of market dynamics. While these methods have been applied in many fields including epidemiology, high-energy physics, and non-equilibrium systems, they have yet to be used to calibrate a market simulator to real market data \cite{brehmer2021simulation,gonccalves2020training,stillman2023graph}. Our work addresses this.


\section{Methods}

In this work, we show how simulation-based inference can be used to calibrate an ABM market simulator. We demonstrate our approach using two ABMs, a simple model containing a set of {\it zero-intelligence traders}, which we refer to as the ZI model, and an extension to the Chiarella model \cite{majewski2020co}, which we refer to as the \textit{extended Chiarella model}. These models are described in \autoref{sec:models}. Most calibration methods for market simulators use a collection of summary features known in economics and finance as {\it stylised facts} \cite{samanidou2007agent,vyetrenko2020get,terasvirta2011stylized}. The stylised facts represent commonly observed features generated by market exchanges. In this work, we use stylised facts as an evaluation measure for our calibration framework and instead use the time-series data directly to train our calibration networks. We describe these summarising features as well as the stylised facts used in \autoref{sec:summstats}. Having outlined both models and summary statistics, we next describe our method for calibration in \autoref{sec:cali}. We use neural posterior estimation (NPE) to estimate the posterior probability distribution of parameter sets, conditioned on an observation. To do so, we use two different types of normalising flows, neural spline flows (NSF) and masked auto-regressive flows (MAFs) to approximate the posterior without requiring the likelihood to be calculated. In order to reduce the dimensions of the summary data, we use an embedding network to transform the data into a low-dimensional feature space. Here, we use a simple multi-layer perceptron (MLP) model as our embedding network. However, as we discuss in \autoref{sec:discuss}, other more complex neural network architectures could also act as an effective embedding networks. This is especially true for architectures that match the inductive biases inherent to market dynamics such as a recurrent or graph neural network. 

\subsection{Models}\label{sec:models}
In this work, we use two different ABMs to test our calibration routine, the zero-intelligence trader model (ZI) and an extension to the Chiarella model. For further details on both, see \cite{preis2006multi} for the ZI model and \cite{majewski2020co} for the extended Chiarella. Future work will look at how this approach performs with other agent-based approaches to market simulation such as using the ABIDES framework \cite{byrd2019abides}.

\subsubsection{Zero Intelligence Trader}

In order to simulate the exchange, we assume that trading is driven by a continuous double auction (CDA) between agents. At each timestep in the simulation, traders can place a limit order or market order in the exchange. Those orders that are not immediately executed are stored in the LOB, where queued limit orders can be cancelled at the beginning of each timestep. Submitted orders are processed sequentially, unmatched orders are queued on the LOB, while orders that can be matched are completely or partially filled. Expired orders are removed. This process is the same for both ZI and extended Chiarella model. 

The ZI model includes, as $N_a$ agents, a set of zero intelligence traders who exhibit non-adaptive and stationary behaviours. These agents trade stochastically, submitting limit orders at each timestep with probability $\alpha$ at a depth $D$ from the mid-price, market orders with probability $\mu$, where $\alpha$ and $\mu$ are calculated as the average number of limit and market orders submitted per timestep, divided by the number of agents, $N_a$. Queued orders are cancelled with probability $\delta$ \cite{preis2006multi}. We note that choice of depth, $D$, is sampled from an exponential distribution from mid-price such that limit orders never cross the spread completely or result in a fill. The price at which the $i^\mathrm{th}$ limit order at time $t$ is submitted is then given by
\begin{align}
p^b_{i,t} &= p^m_t - D_{i,t}, \\
p^a_{i,t} &= p^m_t + D_{i,t},
\end{align}

where $p^b, p^a$ denote the bid and ask price, respectively, $p^m$ is the midprice, and $D$ is the depth, $D \sim \exp(\lambda)$. The volume, $V_{i,t}$, is fixed at 1 and we assume that expiry is end-of-day. We note that the number of limit orders from $N_a$ traders at each timestep follows a Binomial distribution. The number of cancelled orders also follows a Binomial distribution where the expected number of cancelled orders is $\delta N_t$, and $N_t$ is the size of the LOB at time $t$. 

The ZI model provides a baseline to explore market behaviours and the fidelity of market simulation. This model can reproduce stylised facts observed in historical market data despite the relatively simple agent behaviours. However, it is also more constrained in the number of scenarios it is able to reproduce due to its simplicity. We next describe an extension to the Chiarella model which increases the complexity of trader behaviours. 

\begin{figure*}[!t]
    \begin{subfigure}{0.33\textwidth}
        \includegraphics[width=6cm]{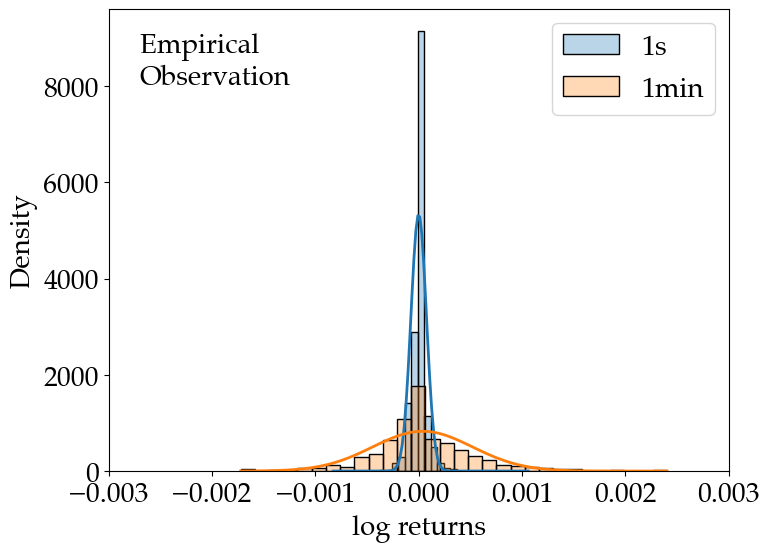}
        \caption{Distribution of (empirical) log returns}
    \end{subfigure}
    \begin{subfigure}{0.33\textwidth}
        \includegraphics[width=5.9cm]{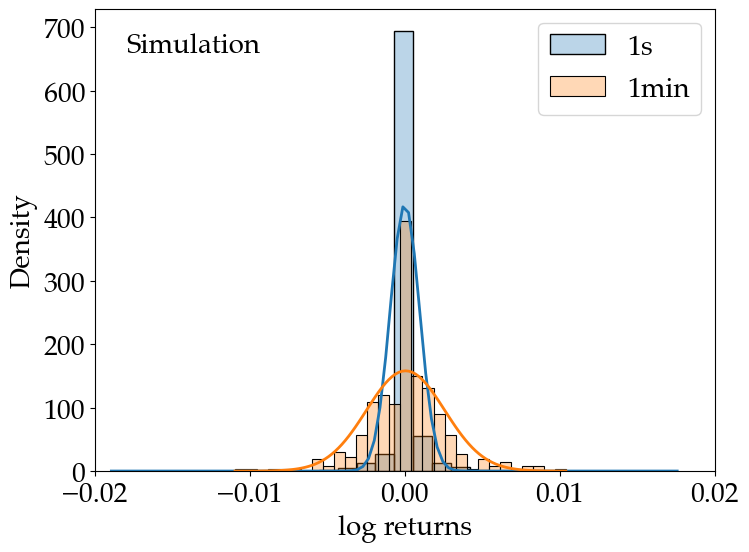}
        \caption{Distribution of (simulated) log returns}
    \end{subfigure}
    \begin{subfigure}{0.33\textwidth}
        \includegraphics[width=5.8cm]{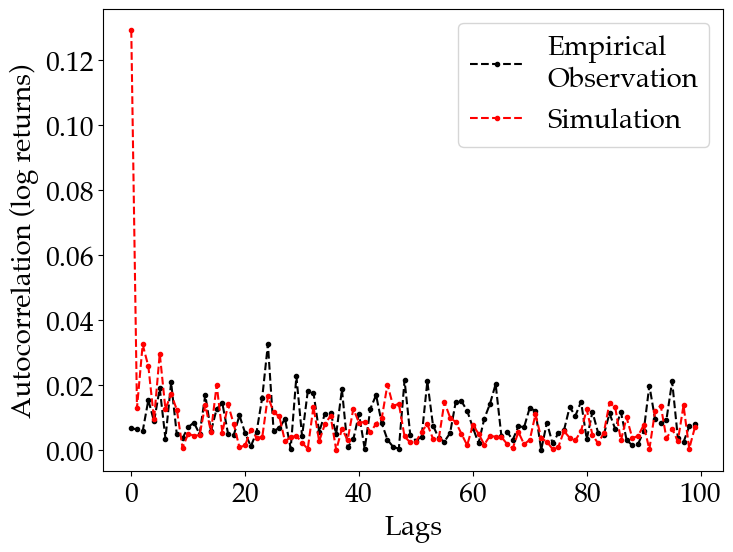}
        \caption{Auto-correlation of log returns}
    \end{subfigure}
    \newline
    \begin{subfigure}{0.33\textwidth}
        \includegraphics[width=6cm]{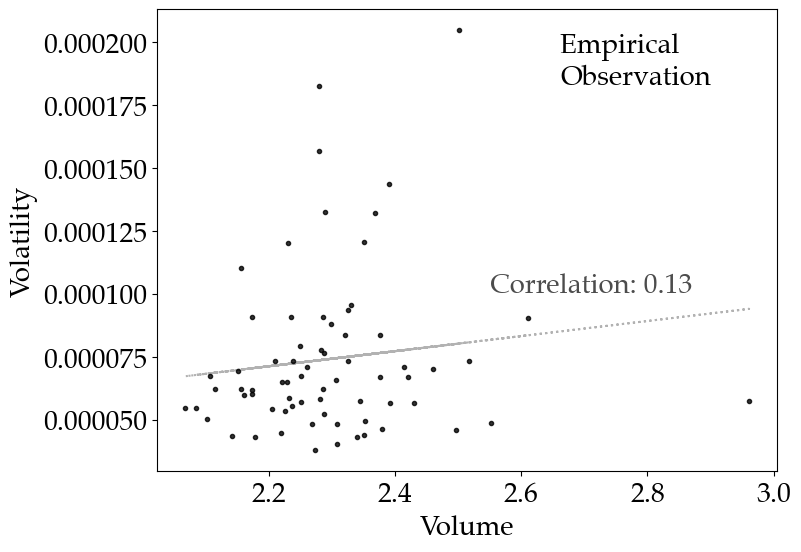}
        \caption{Volume-volatility correlation (empirical)}
    \end{subfigure}
    \begin{subfigure}{0.33\textwidth}
        \includegraphics[width=5.9cm]{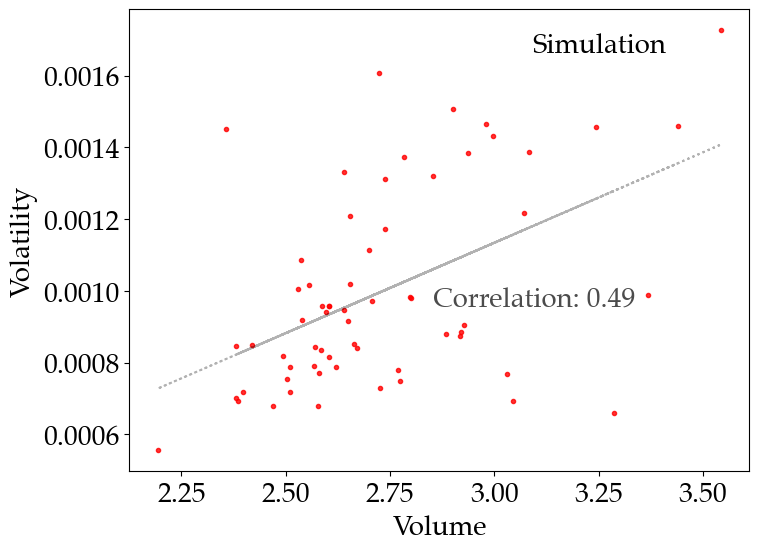}
        \caption{Volume-volatility correlation (simulated)}
    \end{subfigure}
    \begin{subfigure}{0.33\textwidth}
        \includegraphics[width=5.8cm]{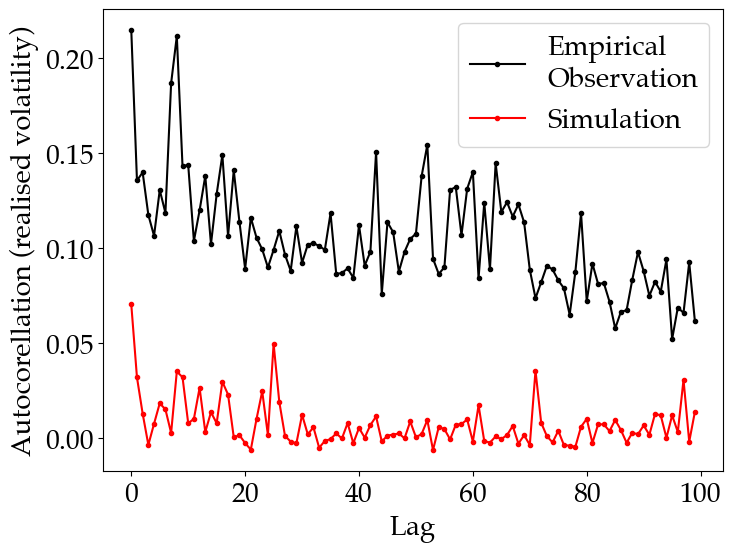}
        \caption{Auto-correlation of realised volatility}
    \end{subfigure}
    \caption{Comparison of stylised facts for simulated and empirical observations}
    \label{fig:stylised}
\end{figure*}

\subsubsection{Extended Chiarella}

The original Chiarella model increased the complexity of trader behaviours by introducing two broad trader classes, \textit{fundamentalists} and \textit{chartists} \cite{chiarella1992dynamics}. Fundamentalists trade based on rational expectations of the underlying fundamental value of the asset in question. On the other hand, chartists respond specifically to price trends such as the momentum of the asset price. This model has been further extended to include a third trader class, \textit{noise traders} \cite{majewski2020co}. The agent behaviours and their relation to the underlying price are described in detail below. 

We first assume that the price dynamics are described by a linear price-impact mechanism, as in \cite{majewski2020co}, 

\begin{equation}\label{eq:chairella}
p_{t + \delta t} - p_t = \lambda d(t, t + \delta t)
\end{equation}
where $\delta t$ denotes simulation timestep, $\lambda$ relates to the overall liquidity in the market and is a first-order approximation of market price sensitivity to demand and supply, also known as `Kyle's lambda', and $d(t, t+\delta t)$ is the aggregated demand in the market due to the trading strategies of various market participants. These market participants are the three class of agents described above; fundamental, momentum (chartist) and noise traders. 

Fundamental traders submit orders based on their perception of the underlying value of the asset, denoted by $v_t$. Given this, they will tend to buy the asset when it is underpriced, namely when $v_t - p_t < 0$ and sell the stock otherwise. We follow the convention in \cite{chiarella1992dynamics} that the aggregated demand of fundamental traders approximates the degree of mispricing within the market, which we denote by $\kappa$. Hence, the aggregated demand for fundamental traders is given by
\begin{equation}
d_\text{fundamental} \sim \kappa(v_t - p_t),
\end{equation}
and where $v_t$ is a pre-assigned exogenous signal. 

Momentum traders instead decide on the price at which to place an order according to the observed price dynamics in the exchange. We assume that these traders follow a momentum strategy, i.e. the upward or downward trend of the asset, where we use an exponentially weighted moving average of past returns as the price trend signal. This trend signal is given by 
\begin{equation}
M_t = (1-\alpha_m)M_{t-1}  + \alpha(p_t - p_{t-1}),
\end{equation}
where $\alpha_m$ is the decay rate. We choose the demand function for momentum traders to be 
\begin{equation}
d_\text{momentum} = \beta \tanh (\gamma_m M_t),
\end{equation}
where $\beta$ controls the overall demand generated by momentum traders, $\gamma_m$ is the saturation of momentum trader demand for very large signals, and where we choose the demand function such that it is monotonic.

Finally, noise traders are designed to capture the behaviours of market participants that are not included in either of the above descriptions. The cumulative demand for these traders is described by a random walk, where overall demand is sampled from a normal distribution with zero mean and standard deviation $\sigma_N$. Hence, $d_\text{noise} \sim \mathcal{N}(0, \sigma_N)$. The overall demand dynamics then evolve according to 
\begin{align}\label{eq:chiarella}
d(t, t + \delta t) &= \kappa(v_t - p_t)\delta t + \beta \tanh (\gamma_m M_t) + \sigma_N \epsilon_t \sqrt{\delta t}\\
dM_t &= -\alpha_m M_t \delta t + \alpha_m dP_t
\end{align}
where $\epsilon_t$ follows a standard normal distribution and where the price dynamics are updated according to \autoref{eq:chiarella}. 


\subsection{Summarising Features}\label{sec:summstats}

In order to evaluate the fidelity of a market simulator, a set of stylised facts are used which denote market features typically observed in historical data. In this work, we use these stylised facts as evaluation metrics to check that our simulator is able to faithfully reproduce realistic market behaviours. These stylised facts are typically extracted from summarising statistics of the data such as the auto-correlation of returns or distribution specific order types. As such, they represent biased metrics for calibrating a simulator to historical data. In order to make our calibration approach unbiased, we use the simulation and historical data directly, without constructing hand-crafted features. To do so, we pass the time-series data through an embedding network which transforms the high dimensional data to low dimensional summary features. We describe both stylised facts and embedding network features below.  

\subsubsection{Stylised Facts}

Stylised facts represent features of markets that are thought to be universally true, regardless of the exchange or asset being traded. These include intuitively true features such as the inability to predict whether a price will go up or down using previous price trends alone. However, further analysis has shown that a number of these stylised facts do not hold under all circumstances, hence reference to them as `stylised'. Despite this, they have become broadly recognised as suitable metrics for assessing the fidelity of a market simulator. The stylised facts that we use for our assessment are described in \cite{bouchaud2001more,terasvirta2011stylized,cont2001empirical,shi2023neural}. These are 
{\it intermittency}, {\it absence of auto-correlations in return series}, {\it concavity of price impact}, {\it gain$/$loss asymmetry in returns}, {\it heavy tails and normality of log returns}, {\it long range memory of absolute returns}, {\it long range dependence of absolute returns}, {\it positive correlation between volume and volatility}, {\it negative correlation between returns and volatility}, {\it volatility clustering}, {\it Gamma distribution in order book volumes}.

\subsubsection{Embedding Network Features}
The stylised facts above are a subset of a larger proportion of known metrics for quantifying the realism of a market simulator. However, their relevance to reproducing the features of a specific trading day is unclear. In this work, we explore new method for summarising market data by training an embedding network to transform high-dimensional data into low dimensional features that are then used for calibration. Here we use a multi-layer perceptron to perform the embedding, with further architecture details given in \autoref{sec:experiments}. 

We use two different data features, the combination of price and total volume at the best bid and ask order at one second intervals, and the volume weighted average price (VWAP) of bid and ask orders. Future work will consider using larger models and more simulation output such as the entire LOB, as we discuss in \autoref{sec:discuss}. 

\subsection{Calibration of Market Model Parameters}\label{sec:cali}
We use simulation-based inference in order to infer the parameter sets which most accurately match the features constructed by the embedding network. In order to verify that our simulation output remains realistic, we also compare our results against the stylised facts but do not use these data to train our networks. Simulation-based inference is a means of inferring the posterior probability distribution of a set of parameters without computing the likelihood, which is typically analytically intractable. It includes methods that use neural networks to perform this inference. We describe this in more detail below. 

\begin{figure*}[!t]
  \begin{subfigure}{0.245\textwidth}
    \includegraphics[width=\textwidth]{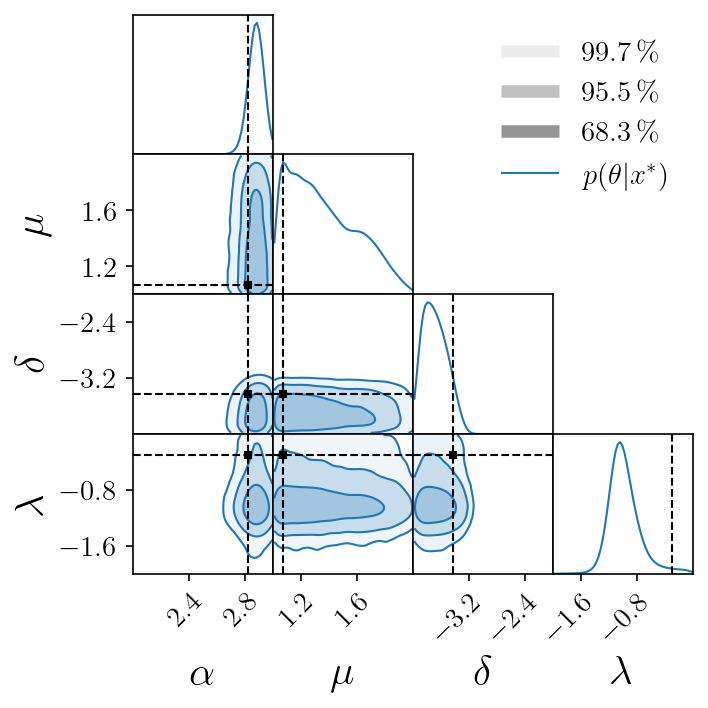} 
    \caption{ZI trader (touch)} 
  \end{subfigure}
  \begin{subfigure}{0.245\textwidth}
    \includegraphics[width=\textwidth]{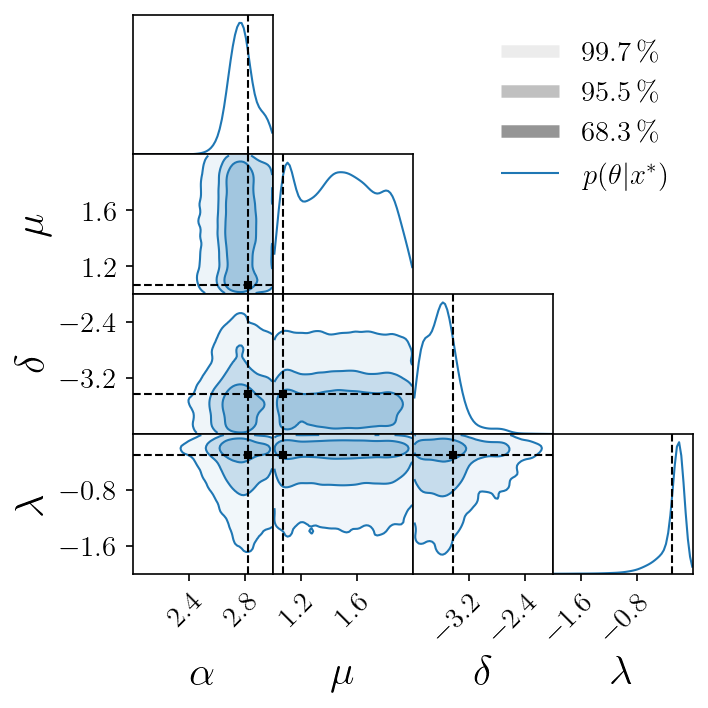} 
    \caption{ZI trader (VWAP)} 
  \end{subfigure}
  \begin{subfigure}{0.245\textwidth}
    \includegraphics[width=\textwidth]{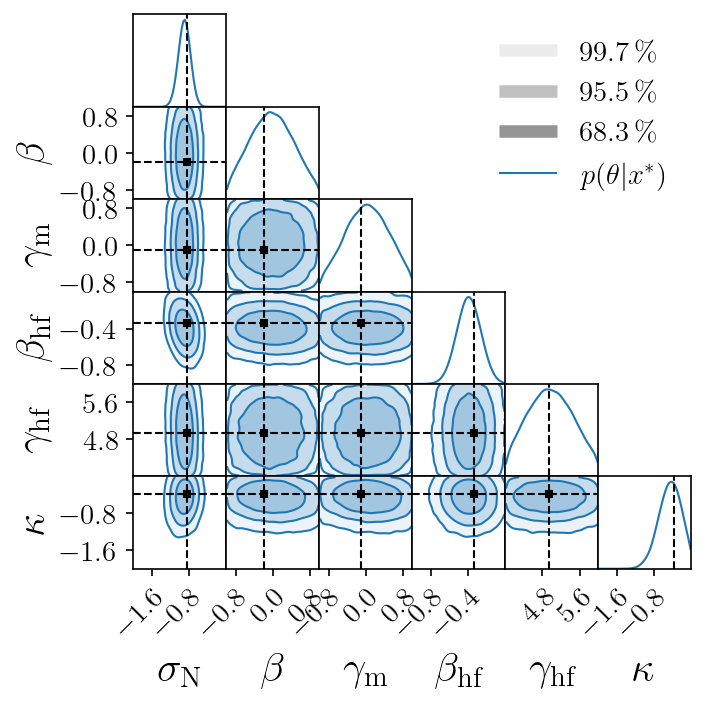} 
    \caption{Extended Chiarella (touch)} 
  \end{subfigure}
  \begin{subfigure}{0.245\textwidth}
    \includegraphics[width=\textwidth]{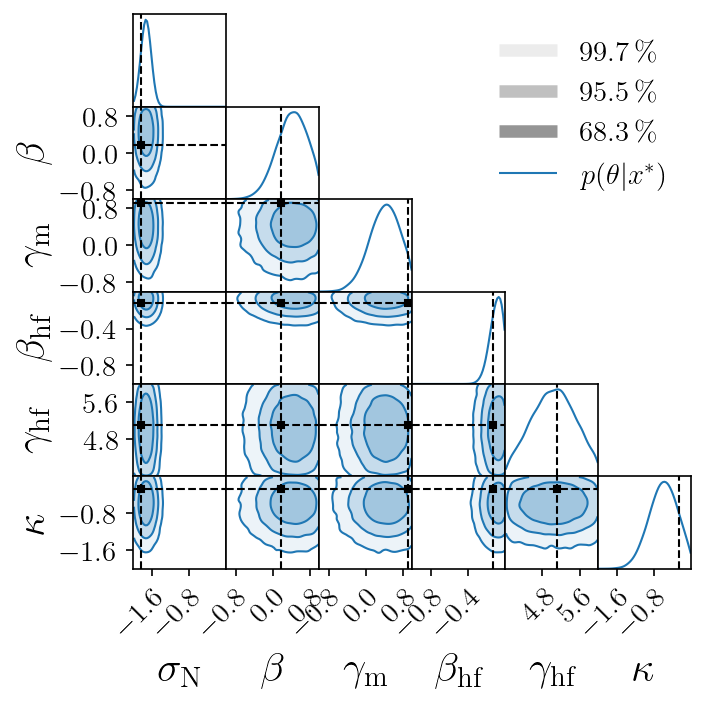} 
    \caption{Extended Chiarella (VWAP)} 
  \end{subfigure}
  \caption{Four calculated posteriors}
  \label{fig:posteriors}
\end{figure*}
\subsubsection{Simulation-based Inference}
Calibration methods often aim to find the point estimate for a parameter set, $\theta^{*}$, that best match historical data, minimising the difference between simulation output, $\mathcal{S}(\theta^*)$, and observations, $\mathbf{x^*}$. Namely, these calibration methods attempt to identify parameter set $\theta^{*}$ such that $|\mathcal{S}(\theta^*) - \mathbf{x^*}| \ll \epsilon$, for some error tolerance $\epsilon$. Simulation-based inference instead seeks to determine the parameter sets that have the highest probability of reproducing observational data, conditional on an observation. This is equivalent finding the posterior distribution of parameters which can be calculated using Bayes' rule, namely
\begin{equation}\label{eq:bayes}
P(\theta|\mathbf{x^*}) = \frac{P(\mathbf{x^*}|\theta)P(\theta)}{P(\mathbf{x^*})},
\end{equation}
where $P(\mathbf{x^*}|\theta)$ is the likelihood (typically intractable for computer simulations), $P(\theta)$ is the initial prior distribution over parameter values, $P(\mathbf{x^*})$ is the evidence of observation, and $P(\theta|\mathbf{x^*})$ is the posterior distribution over parameter sets, which we seek to find. It is worth noting that, under this framework, we do not assume uniqueness in the optimal parameter set. This is equivalent to the claim that there may be multiple values for $\theta$ that are able to reproduce the observations equally well. 

Simulation-based, or likelihood-free, methods avoid the need to calculate the likelihood function in \autoref{eq:bayes} by instead sampling from the joint distribution of simulation output and parameters \cite{cranmer2020frontier}. These methods include ABC that, in its simplest form, samples from the joint distribution, $P(\theta, {\bf x}) = P(\theta)P({\bf x})$ whilst keeping only those values that reproduce historical values within some tolerance, $\epsilon$. More recently, approaches that leverage density estimation techniques in deep learning, such as mixture density networks and normalising flows, have been shown to be both more efficient and accurate than ABC methods \cite{cranmer2020frontier}. These methods include neural posterior estimation (NPE), neural likelihood estimation (NLE), and neural ratio estimation (NRE) \cite{greenberg2019automatic,papamakarios2016fast,hermans2020likelihood}. In this work we focus only on NPE. 

\subsubsection{Neural Density Estimators}
Simulation-based inference estimates the posterior distribution of parameters by sampling from the joint distribution of simulation output and parameter sets. To do so, we use amortized variational inference, which converts the problem of approximating a probability density into a more tractable optimisation problem. Namely, we use neural density estimators, where the posterior is the target density that we seek to estimate and the simulator is the source of the training data for the network. In this work we use normalising flows, specifically neural spline flows (NSF) and masked-autoregressive flows (MAFs) \cite{kobyzev2020normalizing,durkan2019neural,papamakarios2017masked}. For further details on neural density estimation, we refer the interested reader to the discussions in \cite{papamakarios2019neural}.

\section{Experiments}\label{sec:experiments}

We use simulation-based inference to estimate the posterior probability distribution, $P(\theta|\mathbf{x})$, of parameter sets, $\theta$, for a market simulation, given an observation of trading activity $\mathbf{x}$. We sample from a model-specific prior distribution with a fixed simulation budget of $10^4$ simulations, where 1,000 are held back for validation and 1,000 for testing. Simulations were generated on a Dell R750 node, with two Intel 6354 CPUs and all inference was performed on a NVIDIA A30 GPU, where compute resources were accessed provided by NVIDIA LaunchPad Experience.

For the ZI model, we are interested in the following four parameters, $\alpha$, $\mu$, $\delta$, and $\lambda$, which are described in more detail in \autoref{sec:models}. We assume a uniform prior distribution, with lower bound of $[2, 1, -4, -2]$ and upper bound of $[3, 2, -2, 0]$ in log-space. For the extended Chiarella model, we aim to construct the posterior over a six dimensional parameter space, $\sigma_N$, $\beta$,$\gamma_m$, $\kappa$, which are described in \autoref{sec:models}, as well as $\beta_\text{hf}$, $\gamma_\text{hf}$ which represents momentum traders responding to high frequency movements in price. We again assume a uniform prior distribution, with lower bound $[-2, -1, -1, -1, 4, -2]$ and upper bound $[ 0,  1,  1,  0, 6,  0]$ in log-space. For both models, we compare our results to historical data from one days trading of a futures contract on the HKEX exchange. Both prior distributions were selected based on previous data analysis. 

We use the same neural network architecture for both ZI trader model and extended Chiarella: we use four layers for the embedding network with hidden size of 64 and out dimension of 256, and three layers with hidden size of 128 for the neural density estimator. The batch size was 128 and we used a learning rate of 0.001 with plateau scheduler, drop-out rate of 0.1, and early stopping. On average we found training took approximately 80 epochs before converging. We conducted hyper-optimisation using a coarse grid search and fixed the model architectures based on the lowest average test loss for both models. We found that the choice in normalising flow was the only significant factor in differentiating performance across architectures used on either simulation model. 

\subsection{Zero Intelligence Trader}

We find that we are able to effectively recover the market simulator parameters using simulation-based inference. Given the posterior, we are able to identify the four parameters with RMSE of 1.85 (+/- 0.52), when using the mid-price and total volume, and RMSE of 1.21 (+/-0.48) when using the VWAP at the fist level of the LOB. These results are shown in \autoref{fig:posteriors}. We note that the parameter controlling the arrival time for market orders, $\mu$, has a significant uncertainty that spans the prior range and, when using VWAP to estimate posteriors, a distinct bi-modality. This highlights that the ZI model of markets is weakly constrained by arrival time for messages and, furthermore, multiple parameter results may be able to reproduce similar market data. 

Additional to this, we note that the validation and test loss is reduced when using NSF compared to MAF as the neural density estimator. This is expected, given the bi-modality observed in $\mu$, as NSFs are typically more flexible at reproducing complex distributions. We find that using VWAP data with NSF we are able to reproduce the distinct bimodality in $\mu$ along with being more accurate at predicting the other parameter values. It also indicates that $\lambda$, which controls the depth at which orders are submitted, can have a long tail, highlighting the difficulty in determining this parameter value. 

\subsection{Extended Chiarella}
We find that we are also able to identify the parameters for the extended Chiarella model. This is significant, as this model is more complex with an increase in the number of parameters and with greater flexibility in the overall output of the simulator. We find that we are able to calibrate the 6 parameters with a RMSE of 0.77 (+/- 0.26) when using the midprice and total volume data and RMSE of 0.82 (+/- 0.29) when using the VWAP at the first level. While the overall accuracy slightly decreases when using VWAP, we find that the overall uncertainty reduces around those parameters where the posterior probability distribution is already constrained, such as $\sigma_N$, $\beta_{hf}$ and $\kappa$. This indicates a tradeoff on whether the single point estimate for the parameter values is used or the explicit posterior is of interest for example, to evaluate model drift as we discuss in \autoref{sec:discuss}. Moreover, examination of the posterior distribution for specific parameters gives additional insight into model assumptions. 

We observed two interesting features in the estimated posterior probability distribution for the extended Chiarella model. First, as discussed above, we find that a subset of parameters have low variance, such that uncertainty is reduced. These are the parameters controlling the noise traders and the fundamental traders, $\sigma_N$ and $\kappa$. However, for momentum traders, we are able to calibrate the decay term for both high frequency and medium frequency parameters, $\beta$ and $\beta_{hf}$ with low uncertainty. In general, we find that the parameters that control this trader behaviour typically have uncertainties that span the prior. We discuss the implications for this in \autoref{sec:discuss}. 

We note that we were unable to achieve good test performance when using NSFs and were required instead to use MAFs. This may be due to the increased dimension in parameter space, which caused the neural network to over-fit. We trialled different initial learning and dropout rates but observed significant divergence between training and validation loss during training time. Given that MAFs are typically more expressive than NSFs, and are able to capture more complex dependencies between variables, this may be due to the increased dimensionality of the size of the parameter set. As such, we may need a larger simulation budget in order to better estimate the distribution. Future work will investigate this further. 

\subsection{Historical Data}\label{sec:hist}

Having demonstrated that we are able to calibrate synthetic data using neural density estimators and embedding networks, we next use our calibration procedure to identify model parameters specific to a single day of trading. We use data from the Hong Kong exchange (HKEX) which reflects a standard trading day. We first evaluate the stylised facts on the historical data to see which are supported and which are violated. As shown in \autoref{fig:stylised}, we see that (a) log returns follow a typical normal distribution at medium timescales (minutes) but that this departs from normality as we shorten the timescale (seconds), resulting an increase in kurtosis. Additionally, when calculating the skewness at both timescales, we observe a slight asymmetry (0.44). 

Shown in \autoref{fig:stylised}(c), we observe an absence of autocorrelation in the return series, in (d), a positive correlation between volume and volatility and in (e) we see significant volatility clustering at high lag number. We also observe intermittency in historical data, a large Hurst exponent (0.8) and that the autocorrelation of absolute returns rapidly converges to zero reflecting that the historical data has no long range memory or dependencies, and the order book volumes approximate a Gamma distribution, where $\gamma = 0.014, 0.018$ for bid and ask orders, respectively. However, we also note that some stylised facts are not observed in this data, such as a negative correlation between returns and volatility, and significant concavity in the price impact function, which is essentially flat at 0.07. 

\begin{figure}[!t]
  \includegraphics[height=8cm]{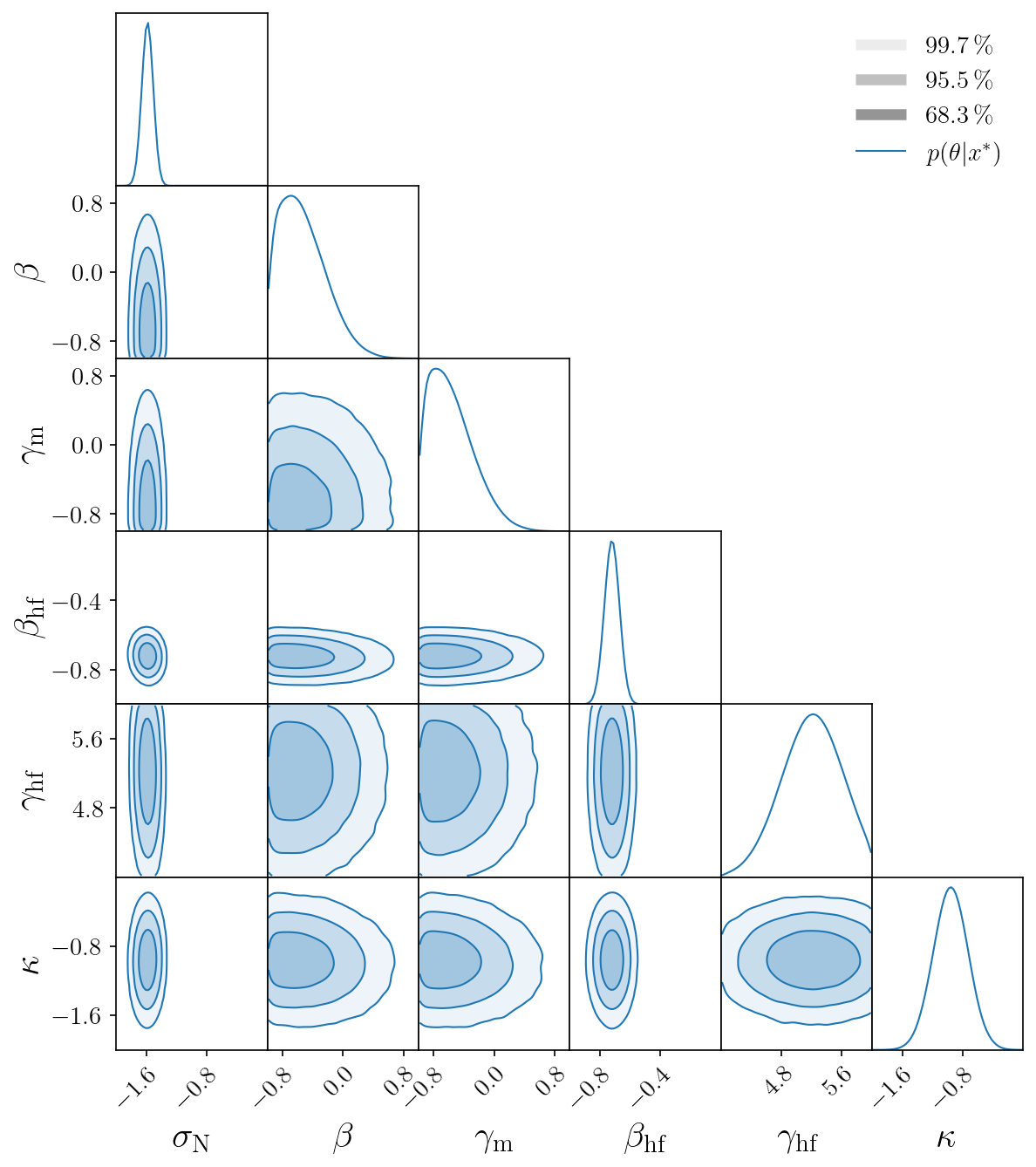}
  \caption{Posterior for extended Chiarella model}\label{fig:posterior_ch}
\end{figure}

In \autoref{fig:posterior_ch}, we show the estimated posterior distribution for the historical data using the VWAP from the fist level of the LOB. We again observe that parameters for the fundamental trader and noise trader are constrained, whereas those for the momentum trader have high uncertainty. Interestingly, we see that the decay rate for high frequency traders has reduced uncertainty, indicating that high frequency behaviours may be significant in this trading day. Future work will investigate this. Shown in \autoref{fig:stylised}, we are again able to reproduce several of the stylised facts, including (a) the heavy tails and normality of log returns, (c) the absence of auto-correlations in return series, and (e) a strong correlation between volume and volatility, as well as intermittency, no long range memory (where the Hurst exponent is 0.76) or dependencies of absolute returns, and a Gamma distribution in the order book volume (where $\gamma = 0.28$ for both bid and ask orders). Interestingly, we observe that our simulator is able to recreate stylised facts that are not present in the historical data such as negative correlation between returns and volatility and a stronger asymmetry in returns (-0.95). We again observe that the price impact function is approximately flat (0.01). The only stylised fact that is observed less strongly in our simulator is the volatility clustering, shown in \autoref{fig:stylised}(f), which decreases with increasing lag but is not consistently positive. 

\begin{figure}[!h]
  \includegraphics[height=8cm]{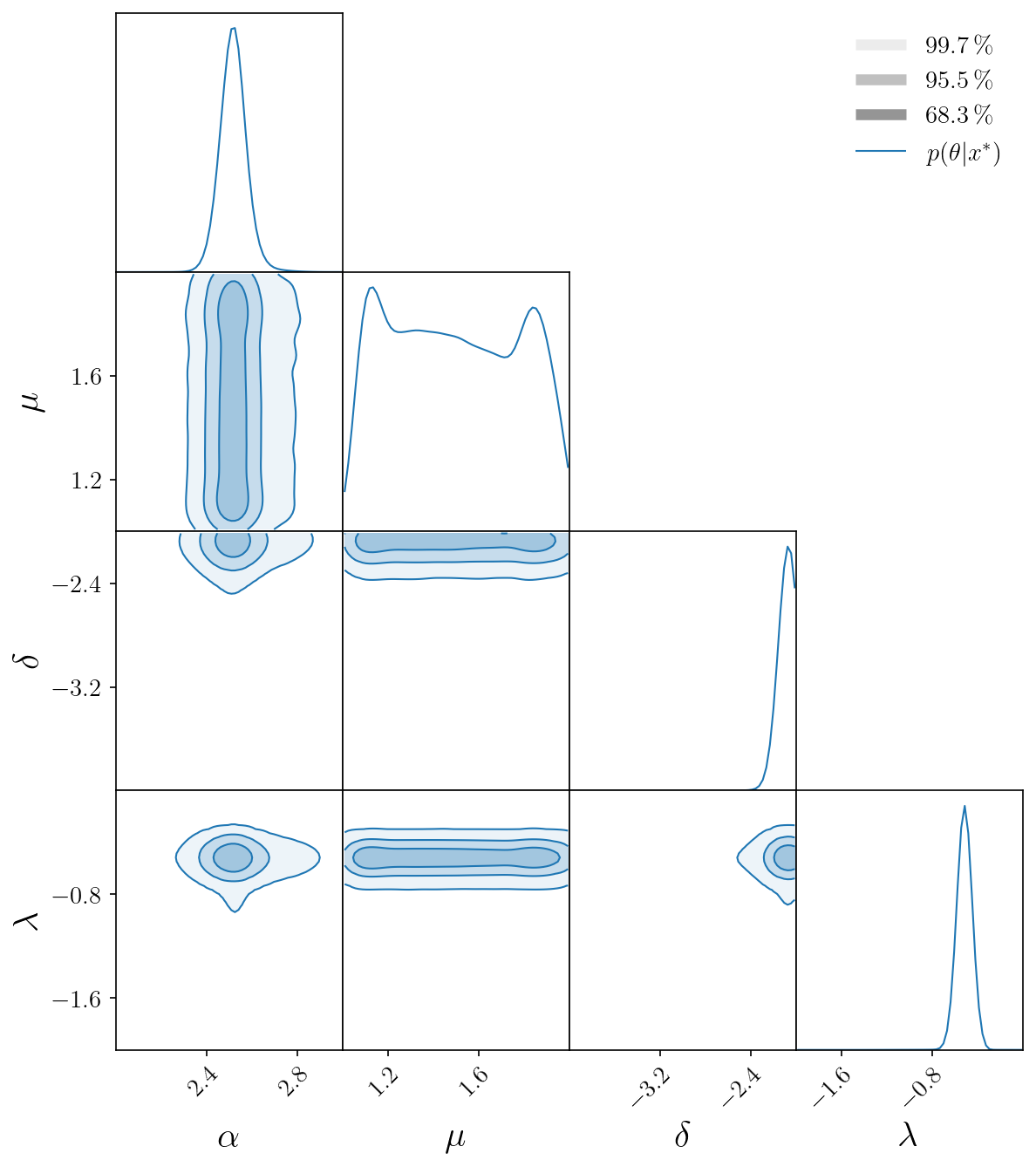}
  \caption{Posterior for ZI Trader model}\label{fig:posterior_zi}
\end{figure}

We next use the historical data to estimate the posterior for the ZI trader model, as shown in \autoref{fig:posterior_zi}. We find that the parameter values are constrained with similar uncertainties as observed when using synthetic data. We again see a sharp bi-modality in the rate at which market orders are submitted that spans the prior. When calculating the stylised facts, we observe the same behaviours as with the extended Chiarella model. However, the price impact function is now convex (with coefficient -0.25) and where there is negligible correlation between returns and volatility (correlation coefficient is -0.0005). 

\section{Discussion \& Future Work}\label{sec:discuss}

We have shown that the combination of neural density estimators with embedding networks is a robust method for inferring parameters of market simulators. We are able to infer the parameters of two different market simulators with high accuracy and, importantly, these market simulators recreate stylised facts common to both simulated and historical data despite the calibration process being blind to these data. This independence from stylised facts is important as the actual market data does not necessarily follow all stylised facts, as we describe in \autoref{sec:hist}. Hence, the determination of which stylised facts are most significant for a single calibration of parameters would require analysis of the underlying observation and an informed choice of weighting across different facts. This is a significant challenge to the scaling of market simulators and introduces a human bias based on which stylised facts to consider. Instead, our approach is able to identify parameter sets that reproduce market features with high probability without reliance on analysis of market data. Furthermore, our approach is amortised, such that the trained density estimator is able to rebuild the posterior probability distribution given historical data without needing to be retrained. This is a significant advantage when seeking to calibrate a market simulator to multiple different days or symbols. 

In this work, we remove the reliance on stylised facts from our calibration procedure by using an embedding network to transform high dimensional simulation output into low dimensional summarising features. While this method is opaque, insofar that it is not clear what the theoretical meaning is of these transformed features, it means that the calibration is data-driven in how it estimates the posterior, without relying on domain knowledge. Here, we use a relatively simple framework for embedding, whereby we convert the mid-price and total volume, or the VWAP at the first level of the LOB, into summary features using a MLP. Future work will explore further improvements to this method by, for example, converting the entire LOB from the day into summary features using neural network architecture with appropriate inductive biases, such as a CNN, as in \cite{zhang2019deeplob}, or GNN, as in \cite{chen2022multivariate}. Other work has already demonstrated how GNNs can be used as embedding networks when calibrating agent-based models, as in \cite{dyer2022calibrating}. However, future work should consider how this approach can be extended to systems where agent dynamics are unobservable.

Finally, we note that this work has implications for model mis-specification and drift in the context of market simulators. First, we note that the posterior is true under the assumption of a correctly constrained prior and that the model is able to describe the underlying data generation process. In a sense, this reflects the distinction between aleatoric and epistemic uncertainty, where we assume that our observation of the market is accurate (if partial) whereas we are unable to quantify the uncertainty associated with the underlying assumptions of the model. Recent work has begun to address these issues, in the context of simulation-based inference, and it would be interesting to investigate this with respect to market simulators \cite{cannon2022investigating,huang2023learning}. Second, and related, we note that this work highlights that parameters can be inferred and monitored as an additional heuristic for monitoring changes in market dynamics. This is especially attractive as such an approach with provide a coarse-grained lens to how approximations in the micro-structure behaviours of trading participants alters, with respect to different stressed conditions, for example. Such an approach would, however, be contingent on a degree of confidence in the underlying model approach and, hence, should likely be first conducted in a `simulation sandbox' where ground truths are available. Finally, we note that our approach combines traditional ABM modelling with deep learning in a post-hoc way, where calibration of market behaviours is performed using deep learning. Hybrid models, which combine both deep and traditional approaches at the level of the generation process, are an exciting new research area that maintains the explainability of traditional methods but with increased flexibility from deep learning models. An open question is how these models are calibrated, especially in an amortised way, given the large number of parameters implicit in neural networks. We look forward to steps taken by the field in the direction of these and other difficult questions in the future. 

\section{Significance}

In this work, we demonstrate that neural density estimators can robustly infer the parameters for a market simulator, based on two distinct theoretical models, a zero intelligence trader, and the extended Chiarella model. We combine neural density estimators to calibrate these models without using stylised facts at calibration time, a significant departure from previous methods for calibration of market simulators. Instead, we use the market simulation data directly, such as the mid-price and total volume, providing an unbiased approach to calibration that can efficiently scale. We identify interesting features of both models from our calibration procedure due to the explicit posterior probability distribution that is calculated around each parameter value, and identify future extensions and frameworks where our work can be used. 

\begin{acks}

We would like to thank the NVIDIA LaunchPad Experience for their help with providing compute resources for this research, especially David Taubenheim, Rafah El-Khatib, Alvin Clark and  Jochen Papenbrock.
\end{acks}

\bibliographystyle{ACM-Reference-Format}
\bibliography{market_sbi}

\end{document}